\pdfoutput=1

\documentclass[11pt]{article}

\usepackage{EMNLP2023}

\usepackage{times}
\usepackage{latexsym}

\usepackage[T1]{fontenc}

\usepackage[utf8]{inputenc}

\usepackage{microtype}

\usepackage{inconsolata}
\usepackage{xcolor}
\usepackage{graphicx}
\usepackage{multirow}
\usepackage{makecell}
\usepackage{subcaption}
\usepackage{amsmath}
\usepackage[ruled,linesnumbered]{algorithm2e}

\title{Reading Between the Tweets: Deciphering Ideological Stances of Interconnected Mixed-Ideology Communities}

\author{
Zihao He, Ashwin Rao, Siyi Guo, Negar Mokhberian, Kristina Lerman \\
USC Information Sciences Institute\\
\texttt{\{zihaoh, mohanrao, siyiguo, nmokhber\}@usc.edu}, \texttt{lerman@isi.edu}
\\
}

\begin{document}
\maketitle
\begin{abstract}
Recent advances in NLP have improved our ability to understand the nuanced worldviews of online communities. Existing research focused on probing ideological stances treats liberals and conservatives as separate groups. However, this fails to account for the nuanced views of the organically formed online communities and the connections between them. In this paper, we study discussions of the 2020 U.S. election on Twitter to identify complex interacting communities. Capitalizing on this interconnectedness, we introduce a novel approach that harnesses message passing when finetuning language models (LMs) to probe the nuanced ideologies of these communities. By comparing the responses generated by LMs and real-world survey results, our method shows higher alignment than existing baselines, highlighting the potential of using LMs in revealing complex ideologies within and across interconnected mixed-ideology communities.\footnote{Code and data are publicly available at \url{https://github.com/zihaohe123/communitylm-message-passing}.}

\end{abstract}

\section{Introduction}


Social media platforms connect people worldwide within digital town squares, transforming how they share information and exchange ideas. However, mass connectivity, has created new vulnerabilities, including rampant misinformation, the formation of echo chambers that confirm people's pre-existing beliefs \cite{cinelli2021echo,rao2022partisan}, and the fragmentation of society into polarized factions that disagree with and distrust each other~\cite{iyengar2019origins}. These developments intensify societal conflicts and undermine trust in democratic institutions \cite{kingzette2021affective,whitt2021tribalism}. 

\begin{figure}[ht]
    \centering
    \includegraphics[width=0.46\textwidth]{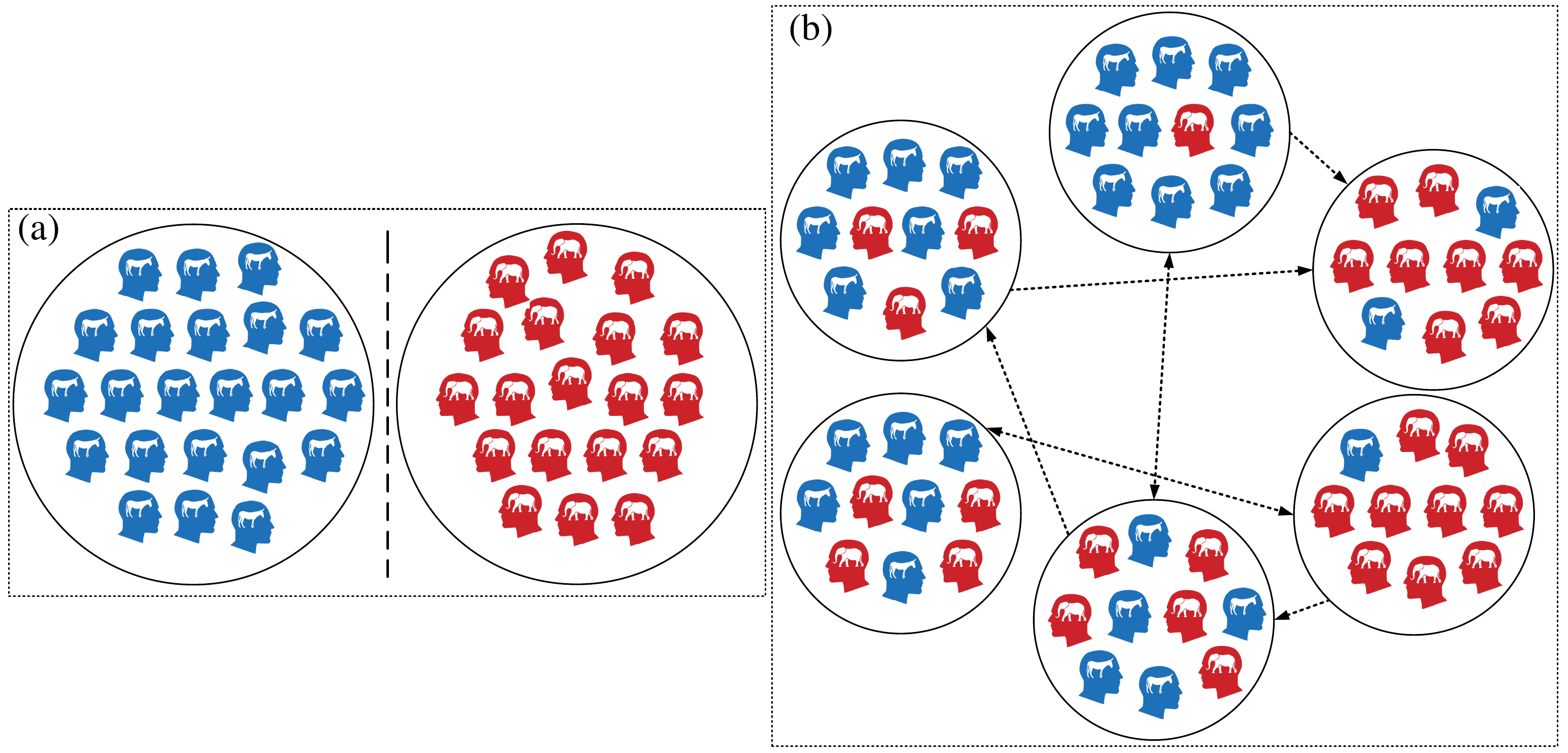}
    \caption{Illustration of online communities, where colors of users represent their political ideologies.
    (a) \textbf{Idealized online communities} that are disconnected and have unified political ideologies. 
    (b) \textbf{Real-world online communities} that are interconnected and have mixed political ideologies covering the full political spectrum. Links between them signify the flow of information and interaction, such as retweeting.
    }
    \label{fig:binary_continuous_communities}
\end{figure}

Given these challenges, understanding the ideological nuances within online communities is essential. Existing works provide insights into political ideologies of online groups~\cite{webson2020undocumented,jiang2022communitylm}; however, they treat ideology as a liberal/conservative binary (Figure \ref{fig:binary_continuous_communities}a) and fail to capture the spectrum of ideologies that may organically emerge in interconnected online communities. 

To bridge this gap, we propose a methodology to uncover interacting communities in political discourse on Twitter that are not merely liberal or conservative, but possess a complex mixture of political ideologies (Figure \ref{fig:binary_continuous_communities}b). To reveal communities' ideological stances, we align GPT-2 language models (LMs) to the language and mindsets of communities by finetuning the models on tweets that the communities generate.
This finetuning, enriched by message passing techniques inspired by Graph Convolutional Networks \cite{kipf2016semi}, leverages the interconnected nature of these communities, allowing for a more robust representation of their ideological stances. With the finetuned LMs, we then probe the stances of the communities towards various targets, including different political figures and social groups, by looking at the sentiment of generated responses. This way we can measure 1) for each target, which communities are more in favor of or against it (target-specific community ranking), and 2) for each community, which targets it favors more and which it is against (community-specific target ranking). 
By comparing the model predicted stances to that from the American National Election Studies (ANES) 2020 Exploratory Testing Survey, our method, when benchmarked against existing baselines, outperforms them on these tasks, validating its effectiveness in capturing the political ideology of interconnected online communities.

Our work highlights the potential of leveraging social media data to reveal the nuanced ideological stances of organically-formed, interconnected online communities. Such insights pave the way for a more informed understanding of the dynamics and shifts in digital attitudes.




\section{Related Work}

\noindent \textbf{Sociolinguistics and Online Communities.}
Existing research examined language change and social dynamics of online communities from a number of perspectives. \citet{danescu2013no} analyzed linguistic change in two online communities of beer enthusiasts, and identified strong patterns within the lifecycle of users within online communities determined by their receptivity to community language norms. \citet{eisenstein2014diffusion} identified geographic differences in the use of language on Twitter and tracked diffusion of linguistic changes across United States, showing that demographically similar communities were more likely to adopt new language norms.


\noindent \textbf{Framing and Ideology.}
Political speech uses framing to make certain aspects of the message salient \cite{lakoff2014all}. By highlighting these aspects, the message can implicitly manipulate the understanding, without explicitly biased argument. Polarized language allows partisans to talk about the same issues using different words to elicit different mental and emotional frames: e.g., talking about ``illegal aliens'' instead of ``undocumented workers'' makes the same group appear threatening \cite{webson2020undocumented}.
\citet{milbauer2021aligning} trained word embeddings on 32 communities from Reddit and discovered multifaceted ideological and worldview characteristics of community pairs, beyond the predetermined ``left'' vs. ``right'' dichotomy of U.S. politics. By using machine translation, \citet{khudabukhsh2021we} studied the political polarization and demonstrated that liberal and conservatives use different expressions as two languages. \citet{he2021detecting} explore the stances of bipartisan news media towards various topics using contextualized word embeddings. Relevant work also showed different patterns of moral framing among liberals and conservatives in the partisan news headlines \cite{mokhberian2020moral} and rhetoric of political elites such as speeches given on the floor of the House and Senate \cite{wang2021moral}.

\noindent \textbf{Probing Community Ideologies with LMs.} 
There is growing interest in aligning language models (LMs) to the ideologies of human communities.
\citet{chu2023language} predicted public opinions from language models by finetuning the models to online news, TV broadcast, and raido shows.
\citet{feng-etal-2023-pretraining} studied politically biased LMs by left and right news and Reddit corpora on hate speech and misinformation detection, and revealed that pretrained LMs reinforce the polarization present in the pretraining corpora.
\citet{jiang2022communitylm} finetuned two language models on tweets from Democratic and Republican communities and probed the ideological stances of the two communities from the models using language prompts that elicit opinions. However, they focus on two manually-defined Democrat/Republican communities and ignore the interactions between them.

\section{Data}
\label{sec:data}

\subsection{ANES Survey}
Following \citet{jiang2022communitylm}, we use the 2020 Exploratory Testing Survey\footnote{https://electionstudies.org/data-center/2020-exploratory-testing-survey} from the American National Election Studies (ANES), which provides ground truth data for evaluating ideological stances predicted by language models. This survey was conducted in April 2020 with a sample of 3,080 US adults. We use the 30 questions from  the \emph{Feeling Thermometers} section, which asked  participants to rate a target---a person or a group---on a scale from 0 to 100. A higher score indicates a warmer, more positive attitude towards the target, and a lower score indicates a cooler, more negative attitude. For each target, the bipartisan ground-truth 
ratings are the average across all scores from liberals and conservatives respectively. Please refer to Appendix \ref{app:anes} for the 30 studied targets.

\subsection{2020 U.S. Election Twitter Data}
We use a public Twitter dataset about the 2020 U.S. presidential election \cite{chen2021election2020}. The data was collected by tracking specific user mentions and accounts tied to the official or personal accounts of candidates, ranging from December 2019 to June 2021. We limit tweets to the time period before April 10 2020, which was the time of the ANES survey we use as ground truth. This way, the dataset does not leak information beyond this date. We filter tweets posted within the U.S.

We identify online communities based on the news co-sharing activities ($\S$\ref{sec:online_communities}). We only keep users with more than 100 followers and those who authored at least one tweet containing a URL to a news article and extract the domain of the URL. The domain represents a news outlet. 
We identify a total of 996 news outlets in this dataset, with the top 10 most shared outlets being \emph{nytimes}, \emph{foxnews}, \emph{washingtonpost}, \emph{cnn}, \emph{breitbart}, \emph{thehill}, \emph{politico}, \emph{nypost}, \emph{cnbc}, \emph{businessinsider}.
After preprocessing, we are left with 41M tweets from 135K users.

\begin{table*}[ht]
\addtolength{\tabcolsep}{-2.0pt}
\centering
\small
\begin{tabular}{ccccl}
\hline
\textbf{comm.} & \textbf{\#users} & \textbf{\#tweets} & \small\textbf{\%lib. tweets} & \multicolumn{1}{c}{\textbf{top-5 shared news outlets}}                                                                          \\ \hline

1  & 38.9K  & 19.3M  & 5  & \textit{\textcolor{red}{foxnews}, \textcolor{red}{breitbart}, \textcolor{red}{nypost}, \textcolor{red}{washingtonexaminer}, \textcolor{red}{wsj}}                                                                    \\

2   & 19.4k  & 3.9M  & 90   & \textit{\textcolor{blue}{nytimes}, \textcolor{blue}{washingtonpost}, \textcolor{blue}{time}, \textcolor{blue}{wapo.st}, \textcolor{blue}{bostonglobe}}                                                                    \\

3  & 15.8k  & 3.9M  & 88 & \textit{thehill, \textcolor{blue}{nbcnews}, \textcolor{blue}{theguardian}, \textcolor{blue}{vox}, \textcolor{blue}{latimes}}                                                                            \\

4  & 11.5K & 2.9M  & 93  & \textit{\textcolor{blue}{rawstory}, \textcolor{blue}{huffpost}, apnews, \textcolor{blue}{thedailybeast}, \textcolor{blue}{politicususa}}                                                                \\

5  & 10.2K & 2.4M  & 89   & \textit{ \textcolor{blue}{politico}, \textcolor{blue}{businessinsider}, \textcolor{blue}{newsweek}, \textcolor{blue}{theatlantic}, bloomberg }                 \\

6  & 7.5K  & 1.5M  & 77   & \textit{\textcolor{blue}{npr.org}, \textcolor{red}{forbes}, reuters, \textcolor{blue}{msn}, bbc}                                                                                     \\

7   & 7.1K  & 1.4M & 92  & \textit{\textcolor{blue}{cnn}, \textcolor{blue}{politico.eu}, \textcolor{blue}{irishtimes}, \textcolor{blue}{baltimoresun}, \textcolor{red}{ccn}}                                                                        \\

8              & 5.2K             & 1.1M              & 87                   & \textit{usatoday, politifact, snopes, factcheck.org, military}                                                                  \\

9   & 3.2K & 0.8M  & 83 & \textit{\textcolor{blue}{abcnews.go}, \textcolor{blue}{markets.businessinsider}, c-span.org,  cs.pn, \textcolor{blue}{sfchronicle}}         \\

10  & 3.0K  & 0.7M  & 30 & \textit{\textcolor{blue}{cnbc}, \textcolor{blue}{nj}, \textcolor{blue}{abc.net.au}, \textcolor{blue}{kansascity}, \textcolor{blue}{mcall}}                                                                                \\

11  & 2.1K  & 0.4M   & 83  & \textit{apple.news, \textcolor{blue}{sun-sentinel}, \textcolor{blue}{seattletimes}, local10, \textcolor{blue}{Salon}}                      \\

12 & 1.8K & 0.3M & 85 & \textit{\textcolor{blue}{abcn.ws}, reut.rs, bbc.co.uk, \textcolor{blue}{sacbee}, \textcolor{blue}{azcentral}}                                                                             \\

13   & 1.3K  & 0.4M  & 38  & \textit{\textcolor{red}{dailymail.co.uk}, \textcolor{red}{spectator.us}, \textcolor{blue}{mercurynews}, thewrap, nejm.org}               \\

14 & 1.2K & 0.3M  & 49  & \textit{\textcolor{blue}{axios}, \textcolor{red}{warroom.org}, \textcolor{red}{bostonherald}, \textcolor{blue}{ajc}, minnesota.cbslocal}                   \\

15  & 1.1K & 0.3M  & 31 & \textit{politi.co, \textcolor{blue}{tampabay}, calmatters.org, \textcolor{red}{fox5ny}, \textcolor{blue}{americamagazine.org}} \\

16 & 1.1K  & 0.3M & 55 & \textit{\textcolor{blue}{cbsnews}, hollywoodreporter, \textcolor{blue}{postandcourier},  modernhealthcare, \textcolor{red}{the-sun}}       \\

17  & 1.0K & 0.2M & 66  & \textit{news.yahoo, \textcolor{red}{christianpost}, \textcolor{blue}{sfgate}, taskandpurpose, mashable}                 \\

18  & 1.0K & 0.2M  & 48 & \textit{\textcolor{red}{reason}, \textcolor{blue}{detroitnews}, \textcolor{blue}{freep}, statnews, mlive}                                                                            \\

19  & 0.8K  & 0.2M  & 96 & \textit{\textcolor{blue}{citylab}, \textcolor{blue}{cbs7}, thestreet, palmbeachpost, \textcolor{blue}{houstonchronicle}}                   \\

20 & 0.5K & 0.1M  & 65   & \textit{\textcolor{blue}{miamiherald}, \textcolor{red}{reviewjournal}, ktla, kvue, on.ktla}                                                                        \\ \hline
\end{tabular}
\caption{Statistics of the 20 largest communities in the \emph{news co-sharing network} of the 2020 Elections Twitter data. Five most popular news outlets are listed for each community.
The liberal and liberal-leaning news outlets are highlighted in blue, and the conservative and conservative-leaning outlets are highlighted in red. Outlets with no overt political bias are shown in black.
}
\label{tab:top-comm-election}
\end{table*}

\section{Exploring Ad-hoc Online Communities}
\label{sec:online_communities}

\subsection{Communities in Co-sharing Network}
\begin{figure}
    \centering
    \includegraphics[width=0.48\textwidth]{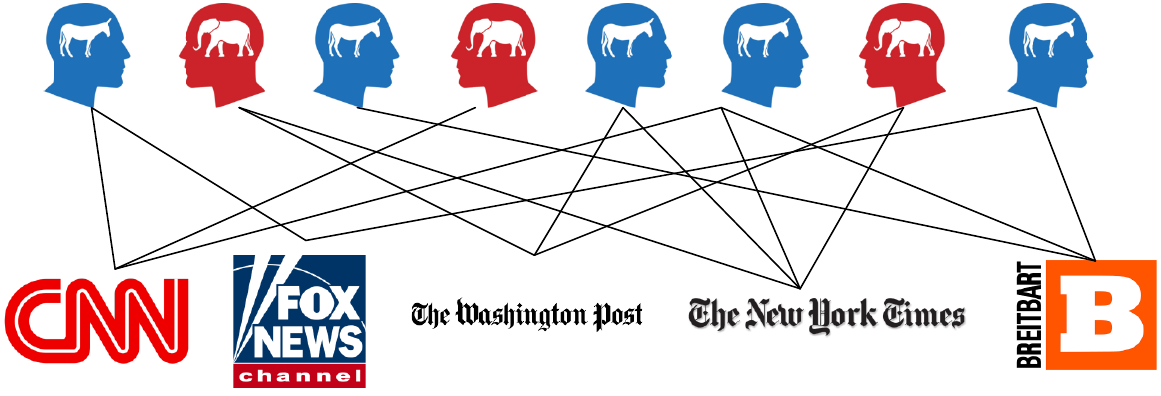}
    \caption{News co-sharing network. A link exists between a user and a news outlet if the user has shared links to articles from the outlet in their tweets. Users having similar news feed are likely from the same online communities. 
    }
    \label{fig:news_cosharing}
\end{figure}

We represent the structure of the information ecosystem as a \emph{news co-sharing network} as shown in Figure \ref{fig:news_cosharing} \cite{faris2017partisanship, mosleh2022measuring, starbird2017examining} and discover communities in it. Utilizing community detection on a \emph{news co-sharing network} is instrumental in discerning the underlying patterns of information dissemination and consumption. By analyzing these communities, we can comprehend how users cluster based on their news-sharing behaviors, offering insights into the sources they prioritize and trust. Such an approach aids in capturing the nuanced dynamics of news engagement, revealing potentially shared interests, regional relevance, or the impact of influential figures. 

We construct a bipartite \emph{news co-sharing network} $G_{co}=(U, V, E)$, where $U$ is the set of users, $V$ the set of news outlets (specified by their domains), and $E$ the weighted edges between them. An edge's weight represents the number of times a user $u$ ($u \in U$) shared links to news stories from this outlet $v$ ($v \in V$) in their tweets.
We use Louvain algorithm \cite{blondel2008fast} to identify communities on $G_{co}$\footnote{We set the resolution to 1, and find that using different resolution values barely change the top 20 detected communities.}. Users that share a similar set of news outlets will be clustered into a community, and each user is only allowed in one community.
As a result, each community $C=(U^{C}, V^{C})$ consists of a set of users $U^{C}$ and news outlets $V^{C}$.
The method identifies 42 communities. We keep the top 20 largest communities, as the users from these communities produce more than 99\% of tweets in the dataset. The statistics and the most shared news outlets in these top 20 communities are shown in Table \ref{tab:top-comm-election}.

\subsection{Mixed Ideologies of Online Communities}
To investigate the ideological leaning of online communities, we first need to identify that of its constituents. Previous works have leveraged on cues in tweet text \cite{rao2021political,cinelli2021echo}, follower relationships \cite{barbera2015birds} and retweet interactions \cite{conover2011political,badawy2018analyzing} to quantify user ideology. In this study, we rely on methods discussed in \cite{rao2021political} to identify user ideology. Specifically, this method extracts ideological cues from tweet text and URLs embedded in them to classify ideology as liberal ($0$) or conservative ($1$).  



Using this approach, we estimate the ideology of users in our presidential election dataset.
Of the 135K users in our sample, we identify 89K as liberals and 45K as conservatives, and the rest users do not have an identified political ideology. The liberal users authored 19M tweets and conservative authored 22M tweets.

For each community, we quantify the fraction of liberal tweets in it in Table \ref{tab:top-comm-election}.
It is important to note that these 20 communities span the  political spectrum, evident by the varying ratios of liberals present within them. This wide range is evident even in the largest, most conservative-leaning community (Community 1) which 
still includes  5\% liberal tweets. More analysis on the ideologies of the communities can be found in Appendix \ref{app:ideo_comm}.

\subsection{Interactions between Online Communities}
Previous works focus on isolated communities, ignoring the interactions between them \cite{jiang2020political, he2021detecting, webson2020undocumented}.
However, retweeting is a popular user activity on Twitter. By retweeting, users endorse the message conveyed in the original tweets \cite{jiang2023retweet, barbera2015birds}. In our dataset, \textasciitilde80\% tweets are either retweets or quoted tweets, and we only focus the former that are more likely to signify endorsement.
Therefore, utilizing messages that have been widely retweeted by a given community helps understand what information the community's members consume, including messages posted by users in other communities. 

To study the interactions between communities, we construct a \textit{community retweet network} among the 20 communities. For a retweet by a user $a$ of a user $b$'s message, we add an edge from the community to which user $a$ belongs to the community where user $b$ is a member. Self-loops are allowed in the network, where a user is retweeting another user in the same community. The edges are weighted, representing the frequency that the retweeting activities happened. For each community, we normalize the weights of its out-edges by its total out-degree. The visualization of the \textit{community retweet network} and more analysis about it are presented in Appendix \ref{app:comm-retweet-net}, where we observe 1) importance of interconnectedness matters, 2) echo chamber phenomenon, 3) diverse news consumption and 4) comparative inclusivity of liberal communities.

\section{Probing Stances of Online Communities}

To study the different opinions and stances of different communities, we delineate each community with a large language model finetuned on this community's corpus. During finetuning, we use the message passing technique to account for the information and opinion shared between communities. Finally, to verify that our models indeed capture communities' political ideology, we test it against multiple baselines on stance prediction toward 30 politically salient entities or groups. The results show the outstanding performance of our method.

\subsection{Methodology}
\noindent \paragraph{Finetuning Language Model.}
\label{sec:model_finetuning}
A community's corpus $D$ consists of tweets made by all users within the community. For each community, we finetune a generative language model GPT-2 \cite{radford2019language} on the corpus using the causal language modeling task. During finetuning, the language model is aligned to the language and mindsets from the community \cite{jiang2022communitylm}. 

\noindent \paragraph{Message Passing between Community Corpora.}
Given the established interconnected nature of communities in the \emph{community retweet network}, it becomes paramount to consider these connections when fine-tuning individual language models for different communities.
Drawing inspirations from Graph Neural Networks (GNNs) where nodes exchange information with their neighbors (message passing), we propose to finetune the community language models using message passing between their corpora. The intuition is that if a community $C_i$ retweets another community $C_j$, then $C_i$ is likely to share similar ideologies as $C_j$ \cite{barbera2015birds}.

We represent the corpus of community $C_i$ as $D_i = (t^i_1, t^i_2, ..., t^i_{|D_i|})$, where each $t^i_k$ denotes a specific tweet in $D_i$. $D_i$ contains the liberal subset $D_i^{lib}$ consisting of liberal tweets and the conservative subset $D_i^{con}$ consisting of conservative tweets. $r^{lib}_i$ and $r^{con}_i$ represent the fractions of liberal and conservative tweets respectively in community $C_i$ and $r^{lib}_i+r^{con}_i=1$.
$N^+(C_i)$ denotes the outgoing neighbors of $C_i$. The normalized edge weight, representing the strength of connection between two communities $C_i$ and $C_j$, is denoted by $w_{ij}$. In the \emph{community retweet network}, $N^+(C_i)$ signifies the communities that have been retweeted by $C_i$. It is important to note that $C_i$ itself can be included in $N^+(C_i)$ as a community can retweet itself.

\begin{figure}
    \centering
    \includegraphics[width=0.46\textwidth]{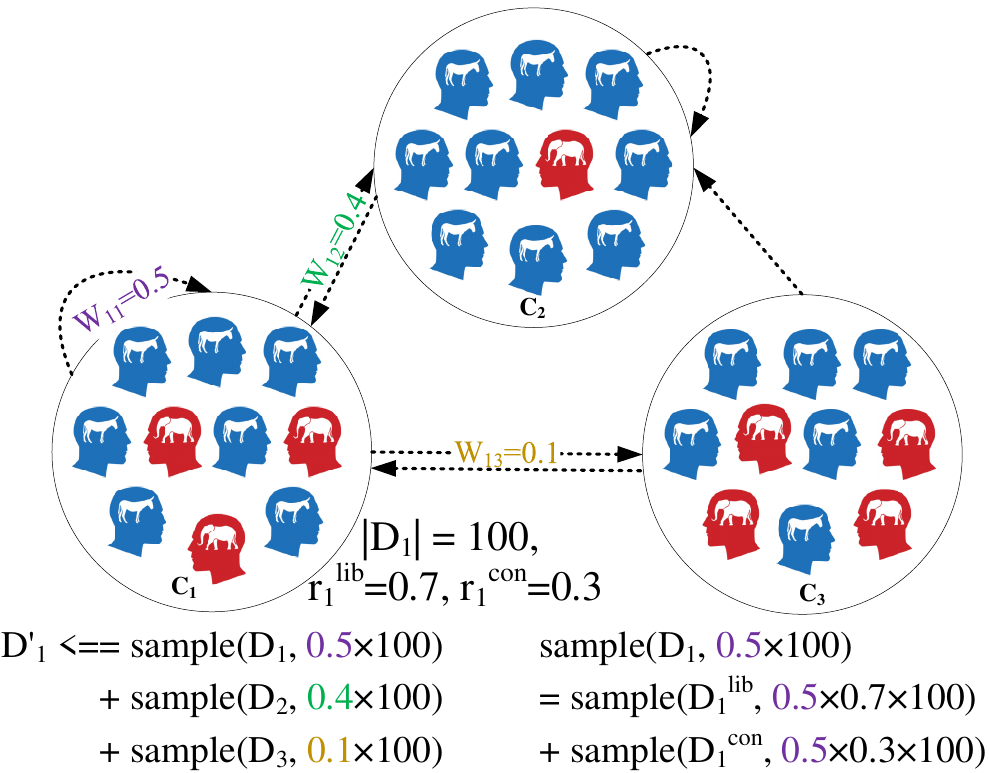}
    \caption{
    Illustration of message passing of community $C_1$ in a simplified retweet network with three communities. 
    The source node of an edge is the retweeting community, and the target node is the retweeted community.
    $D_1$ (the corpus of $C_1$) contains 100 tweets, where the fraction of liberal and conservative tweets are 0.7 and 0.3 respectively. The normalized out degrees for community $C_1$ are shown on its out edges. 
    At each step of message passing, community $C_1$ exchanges information and updates its corpus with its neighboring communities including itself, based on its retweeting activities.
    The numbers of liberal and conservatives tweets sampled from the neighbors are based on the existing ration within $C_1$.
    }
    \label{fig:mp}
\end{figure}

The language model of each community $C_i$ is finetuned on its corresponding corpora $D_i$ over a total of $x$ steps, with message passing performed in intervals of $y$ ($y<x$). 
During message passing, $C_i$ exchanges information with its neighboring communities, while retaining the ratio of liberal and conservative tweets. This is achieved by updating its corpus to $D'_i$:
\begin{align*}
    & D'_i \Leftarrow \sum_{C_j \in N^+(C_i)} \textnormal{sample} (D_j, w_{ij}*|D_i|), \\
    & \textnormal{sample} (D_j, w_{ij}*|D_i|) \\ & = \textnormal{sample} (D_j^{lib}, w_{ij} * r_i^{lib}*|D_i|) \\
    & + \textnormal{sample} (D_j^{con}, w_{ij} * r_i^{con}*|D_i|), 
\end{align*}
where $D_j^{lib}$ and $D_j^{con}$ are the liberal and conservative corpus of $C_j$, and $\textnormal{sample}(D, k)$ represents the corpus of $k$ tweets randomly sampled from $D$. The sum of two corpora implies their merging. Note that the updated corpus $D'_i$ is of the same size as $D_i$. An illustrative example is shown in Figure \ref{fig:mp}.

Utilizing message passing, we ensure that the learning process of one community-specific model benefits from the insights and nuances found in its interconnected neighbors. This approach acknowledges the reality that no community exists in isolation; they frequently influence and are influenced by their surrounding communities. 
In addition, to ensure that the liberal-conservative ratio is preserved within each community, we sample liberal and conservative tweets from neighboring communities based on the existing ratio within each respective community.

This method of using message passing introduces minimal computational overhead and is highly scalable. Notably, it does not necessitate collective fine-tuning of multiple language models, which allows for more flexible and efficient training.

\subsection{Evaluation Protocol}

\noindent \textbf{Community Response Generation.}
For each finetuned community language model, we use four prompts \cite{jiang2022communitylm} to probe its attitude towards a target $X$, which represents  one of 30 politically salient entities or groups (Appendix \ref{app:anes}): (1) ``X'', (2) ``X is/are'', (3) ``X is/are a'', (4) ``X is/are the''. For each target, the model generates $n$ responses using each prompt.

\noindent \textbf{Community Stance Aggregation.}
Following \citet{jiang2022communitylm}, we calculate the sentiment of the response and use it as a proxy of the community's stance towards the target. We use Twitter sentiment classifier \emph{cardiffnlp/roberta-base-sentiment-latest} \cite{barbieri2020tweeteval, loureiro2022timelms} to measure sentiment: negative (-1), neutral (0), or positive (1). The average sentiment score $\hat{s}_{i \rightarrow j}$ over all $n$ generated responses is a measure of community $C_i$'s attitude towards the target $t_j$. Please refer to Appendix \ref{app:stance_detection} for the reasoning behind using sentiment analysis as a proxy of stance detection.

\noindent \textbf{Community Stance Reweighting.} 
The ANES survey reports the liberal rating toward the target $t_j$ (averaged over all liberal participants) as $s^{lib}_j$, and the conservative rating (averaged over all conservative participants) as $s^{con}_j$. 
As we demonstrate in $\S$\ref{sec:online_communities}, every ad-hoc community has a mixed ideology with users from both sides. Thus, delineating the ideology of these communities entails taking into account such mixture of ideologies. As a result, we use the weighted average of the two-sided ratings from the survey by the fractions of liberal tweets and conservative tweets in the community as the ground truth score of a target.
Specifically, we denote the rating (i.e., ground truth stance score) of community $C_i$ towards the target $t_j$ as $s_{i \rightarrow j} = r^{lib}_i * s^{lib}_j + r^{con}_i * s^{con}_j$, where $r^{lib}_i$ and $r^{con}_i$ represent the fractions of liberal and conservative tweets respectively in community $C_i$ and $r^{lib}_i+r^{con}_i=1$.

\begin{figure}
    \centering
    \includegraphics[width=0.48\textwidth]{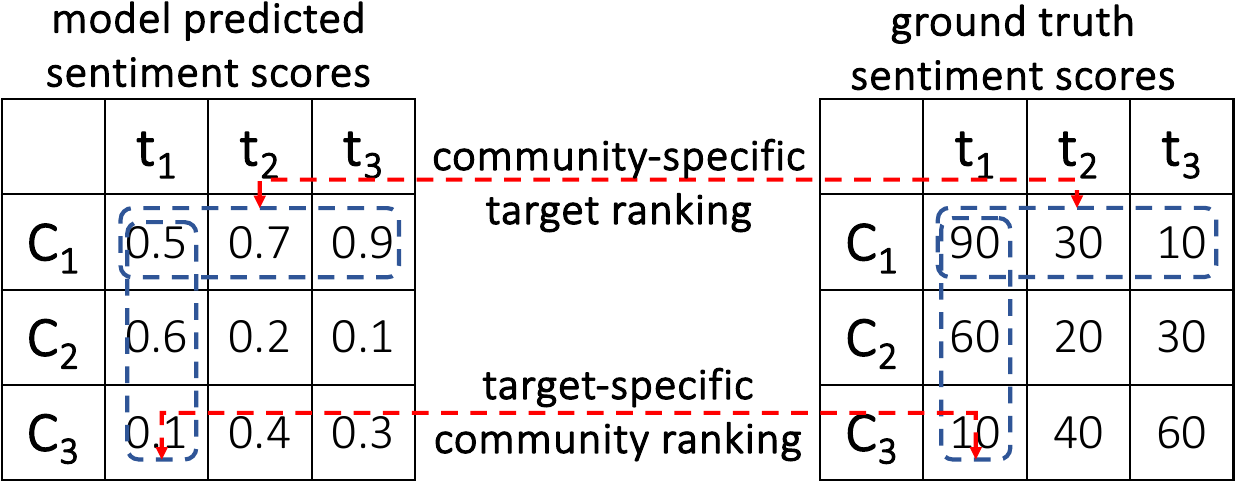}
    \caption{Illustration of target-specific community ranking and community-specific target ranking using a toy example with three communities and three targets.}
    \label{fig:ranking_tasks}
\end{figure}

\noindent \textbf{Target-specific Community Ranking.}
Given a target, we try to capture the stances of different communities towards it, i.e., identify which communities favor the target and which are against it (Figure \ref{fig:ranking_tasks}). 
Specifically, for target $t_j$, we compare two lists of sentiment scores from $N$ communities towards it: one from the model prediction $\hat{S}_{t_j}=\{\hat{s}_{0 \rightarrow j}, \hat{s}_{1 \rightarrow j}, ..., \hat{s}_{N \rightarrow j} \}$, and the other from the reweighted ground truth $S_{t_j}=\{s_{0 \rightarrow j}, s_{1 \rightarrow j}, ..., s_{N \rightarrow j} \}$. The correlation between them is measured by a ranking coefficient $\textnormal{rank\_corr}_{t_j}(\hat{S}_{t_j}, S_{t_j})$, which varies between -1 and 1 with 0 implying no correlation.
The final target-specific community ranking coefficient is averaged over all $M$ targets, as $\frac{1}{M} \sum_{j=1}^{M} \textnormal{rank\_corr}_{t_j}(\hat{S}_{t_j}, S_{t_j})$.

\noindent \textbf{Community-specific Target Ranking.}
Given a community $C_i$, we also want to measure which targets the community favors more and which it is against (Figure \ref{fig:ranking_tasks}). Given two lists of sentiment scores from the language models and reweighted ground truth of community $C_i$ towards $M$ targets, the ranking coefficient between them is $\textnormal{rank\_corr}_{C_i}(\hat{S}_{C_i}, S_{C_i})$.
The final community-specific target ranking coefficient is averaged over all $N$ communities, as $\frac{1}{N} \sum_{i=1}^{N} \textnormal{rank\_corr}_{C_i}(\hat{S}_{C_i}, S_{C_i})$.

\subsection{Baselines}
We compare our finetuned language model with message passing between corpora to the following baselines.

\noindent \textbf{Pretrained GPT-2} \cite{radford2019language}. The vanilla pretrained GPT-2. To align the model to different communities with varying ratios of liberals and conservatives, when generating responses we append a context to the prompt: ``As an independent who agrees with Democrats $x$\% percent of the time and Republicans $y$\% percent of the time, I think'' where $x$ and $y$ represent the fractions of liberal and conservative tweets in that community.

\noindent \textbf{Pretrained GPT-3} \cite{brown2020language}. The original GPT-3 Ada.
The same context is used for generating responses as for the pretrained GPT-2. The generations are obtained by querying the API\footnote{The GPT-3 Ada API has been suspended by OpenAI.}. 
We do not use GPT-4 \cite{ouyang2022training} because it refuses to generate personal opinions or beliefs.

\noindent \textbf{Finetuned GPT-2} \cite{jiang2020political}.  GPT-2 finetuned on each community corpus independently, without using interactions between communities by message passing.

\begin{table*}[ht]
\addtolength{\tabcolsep}{-2.0pt}
\centering
\small

\begin{subtable}{\linewidth}
\centering
\begin{tabular}{ccccccccc}
\Xhline{1.0pt}
\multirow{2}{*}{} & \multicolumn{2}{c}{\textbf{Pretrained GPT-3}} & \multicolumn{2}{c}{\textbf{Pretrained GPT-2}} & \multicolumn{2}{c}{\textbf{Finetuned GPT-2}} & \multicolumn{2}{c}{\textbf{Finetuned GPT-2 + MP}} \\ \cline{2-9} 
            & Spearman(\%)               & Kendall(\%)              & Spearman(\%)            & Kendall(\%)          & Spearman(\%)               & Kendall(\%)               & Spearman(\%)                 & Kendall(\%)                  \\
\textbf{P1} & 8.7                 & 6.0                  & 6.6$\pm$1.9            & 4.9$\pm$1.5            & 39.8$\pm$1.3            & 31.6$\pm$1.3           & \textbf{46.7$\pm$1.4}     & \textbf{38.1$\pm$1.1}     \\
\textbf{P2} & -3.1                 & -2.8                & 9.1$\pm$2.7            & 7.2$\pm$1.6            & 41.8$\pm$0.8            & 32.5$\pm$0.5           & \textbf{48.7$\pm$0.7}     & \textbf{39.2$\pm$0.8}     \\
\textbf{P3} & 1.5                   & 1.6                & 1.2$\pm$2.9           & 9.4$\pm$2.5            & 39.8$\pm$0.8            & 30.7$\pm$0.6           & \textbf{48.9$\pm$1.5}     & \textbf{38.8$\pm$1.4}     \\
\textbf{P4} & 6.3                   & 4.8                & 9.3$\pm$2.6            & 7.3$\pm$2.1            & 45.3$\pm$1.0            & 34.9$\pm$0.9           & \textbf{49.8$\pm$0.8}     & \textbf{39.5$\pm$0.7}     \\ \Xhline{1.0pt}
\end{tabular}
\caption{Results on target-specific community ranking. For each target, scores of the 20 communities from the models and the ANES survey are compared. Reported correlations are averaged over all 30 targets.}
\label{tab:target-specific}
\end{subtable}
  
\vspace{0.1cm} 
  
\begin{subtable}{\linewidth}
\centering
\begin{tabular}{ccccccccc}
\Xhline{1.0pt}
\multirow{2}{*}{} & \multicolumn{2}{c}{\textbf{Pretrained GPT-3}} & \multicolumn{2}{c}{\textbf{Pretrained GPT-2}} & \multicolumn{2}{c}{\textbf{Finetuned GPT-2}} & \multicolumn{2}{c}{\textbf{Finetuned GPT-2 + MP}} \\ \cline{2-9} 
            & Spearman(\%)        & Kendall(\%)              & Spearman(\%)              & Kendall(\%)               & Spearman(\%)              & Kendall(\%)              & Spearman(\%)                & Kendall(\%)                 \\
\textbf{P1} & -3.2                   & -2.5                 & -16.7$\pm$0.8            & -9.9$\pm$0.6            & 12.5$\pm$0.3           & \textbf{8.9$\pm$0.2}          & \textbf{13.0$\pm$0.6}     & 8.8$\pm$0.3     \\
\textbf{P2} & -5.8                  & -3.0                & -23.3$\pm$1.2              & -13.6$\pm$1.2            & 6.3$\pm$1.0            & 5.0$\pm$0.6           & \textbf{7.1$\pm$0.6}     & \textbf{5.0$\pm$0.5}     \\
\textbf{P3} & -5.8                   & -4.7                 & -25.3$\pm$1.3            & -15.5$\pm$0.8           & \textbf{14.5$\pm$0.7}            & \textbf{10.3$\pm$0.5}           & 14.0$\pm$0.4     & 10.2$\pm$0.3     \\
\textbf{P4} & -21.1                & -14.3                & -23.4$\pm$0.8            & -14.9$\pm$0.5           & 16.1$\pm$0.5            & 10.4$\pm$0.5           & \textbf{16.1$\pm$0.4}     & \textbf{10.6$\pm$0.3}     \\ \Xhline{1.0pt}
\end{tabular}
\caption{Results on community-specific target ranking. For each community, scores of the 30 targets from the models and the ANES survey are compared. Reported correlations are averaged over the top-10 largest communities.}
\label{tab:community-specific}
\end{subtable}

\addtolength{\tabcolsep}{2.0pt}
\caption{Spearman and Kendall tau rank correlation coefficients on two ranking tasks. 
The coefficients measure the ranking correlation of model's predictions of community's stances towards the targets to the ground truth ranking obtained from the ANES survey. 
P1 through P4 stand for the four prompts used to query the model: (1)“X”, (2)“X is/are”, (3) “X is/are a”, and (4) “X is/are the”. MP stands for message passing.
The best results using different prompts on Spearman correlation and Kendall tau are highlighted in bold.
}

\label{mergedtable}
\end{table*}

\subsection{Experimental Setup}

\noindent \textbf{Tweet Processing.} We removed URLs (after constructing the \emph{news co-sharing network}) from the tweet texts. For tweets that are cut off by an ellipsis due to exceeding the max length in querying the Twitter API, we removed the ellipsis as well as the characters preceding it.

\noindent \textbf{Backend Language Model.} Following \citet{jiang2020political}, we pick GPT-2 as our backend generative language model. We do not use a larger open-sourced language model like Llama \cite{touvron2023llama} for the following reasons. First, our goal is to proactively predict opinions towards people or groups. Therefore, for fair evaluation, the language model should be pretrained on data curated before April 2020 when the ANES survey was conducted. However, recent large language models are pretrained using data after this time. Second, we argue that our method to finetune language models with corpora message passing to probe community ideologies is highly portable and can be used with any backend language model. By demonstrating its effectiveness on GPT-2, we believe that it will generalize to larger language models. For setup of GPT-2 finetuning, please refer to Appendix \ref{app:exp}.


\noindent \textbf{Evaluation.} For a finetuned GPT-2 model on a community, it generates 1,000 responses for a target using each prompt with greedy decoding. We sample the longest 850 responses from them to filter out ones that immediately stops following the prompt.
We run the generations for 5 times with different random seeds. The average performance over different runs are reported. For the GPT-3 Ada model, we only query it once with 1,000 responses due to the cost.
We use Spearman's rank correlation coefficient and Kendall's tau as the metrics for evaluating the two ranking tasks. 

For target-specific community ranking, the reported metrics are averaged over all 30 targets. For community-specific target ranking, they are averaged over the top-10 largest communities, as the 11th ro 20th communities contain fewer than 0.5M tweets, which are insufficient for the models to capture the internal differences between the targets within each community (as demonstrated by the negative correlations by all studied models).

\subsection{Results}

\textbf{The overall results} on target-specific community ranking and community-specific target ranking are shown in Table \ref{tab:target-specific} and \ref{tab:community-specific}. First, for target-specific community ranking, using messaging passing between community corpora (our method) achieves state-of-the-art performance, consistently outperforming all baselines on all prompts; for community-specific target ranking, our method outperforms most baselines.
It is worth noting that in contrast to \citet{jiang2020political}, who use classification task to decide which of the two communities favors a target more, the ranking tasks we use to evaluate performance over multiple communities and targets are much more challenging. 
Second, pretrained GPT-2 and pretrained GPT-3 barely captures any correlation, because they fail to understand the context we provide (``As an independent who agrees with Democrats $x$\% percent of the time and Republicans $y$\% percent of the time, I think'') to align them to communities, demonstrating few differences between different communities. This is expected to a certain degree because these models are not finetuned on instruction-following \cite{ouyang2022training}.
Finally, out of the two ranking tasks, community-specific target ranking  is a harder task, where the model needs to capture the intrinsic differences in attitudes within a community towards the targets. 
This is even more challenging when one community barely mentions the target, providing the language model little information to learn about it. However, our method allows the language model to learn about the target from the neighboring communities which the community retweets. This improves the learned community insights, increasing the correlations compared to the finetuned GPT-2 baseline in most cases.

\begin{figure}[ht]
\centering

\begin{subfigure}{0.4\textwidth}
\centering
    \includegraphics[width=\textwidth]{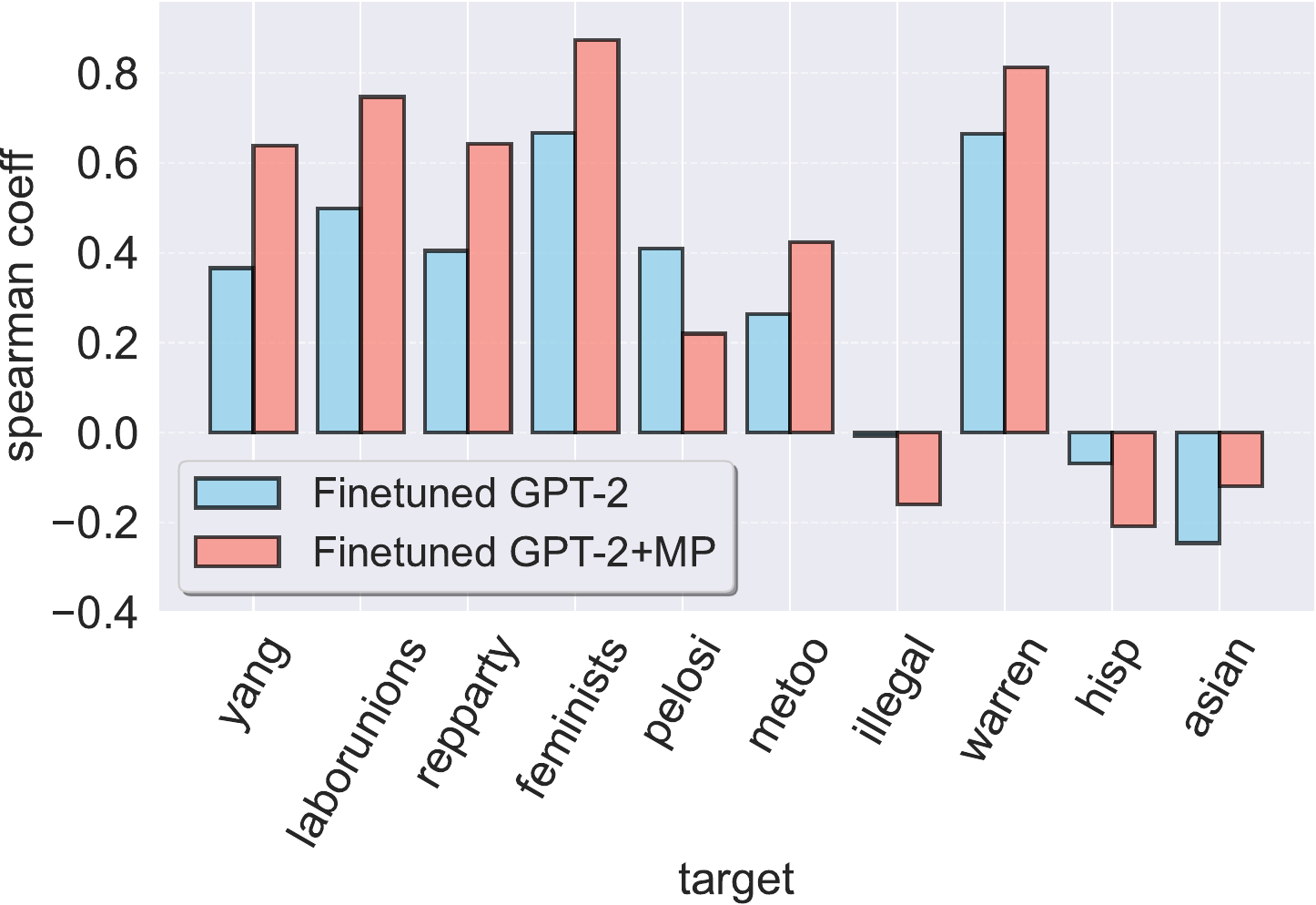}
    \caption{Target-specific community ranking.}
    \label{fig:target-specific-cross-community}
\end{subfigure}
  
\vspace{0cm} 
  
\begin{subfigure}{0.4\textwidth}
    \centering
    \includegraphics[width=\textwidth]{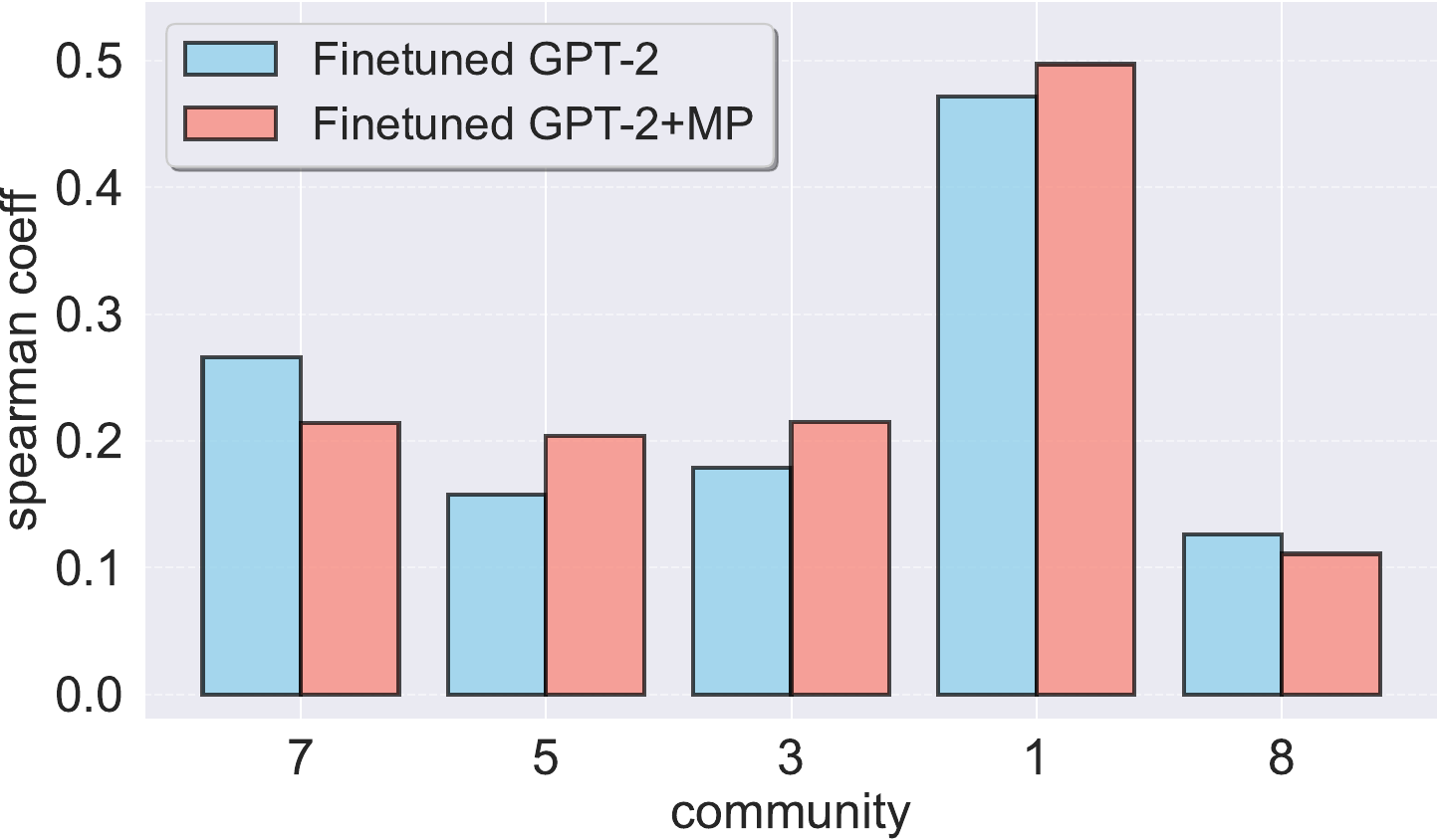}
    \caption{Community-specific target ranking.}
    \label{fig:community-specific-cross-taret}
\end{subfigure}
  
\caption{Spearman's rank correlation coefficients using Prompt 4 (``X is/are the'') for 10 targets and 5 communities of the finetuned GPT-2 baseline and our method on two ranking tasks. The targets/communities are the ones with the largest coefficient change between the two methods, either positively or negatively. From left to right, the targets/communities are sorted by the magnitude of their performance changes.
}
\label{mergedfigure}
\end{figure}

\noindent \textbf{In-depth Analysis.} Figure \ref{fig:target-specific-cross-community} shows the Spearman coefficients with largest differences on target-specific community ranking using Prompt 4 for 10 targets, between the finetuned GPT-2 baseline and our method using message passing.
Similarly, Figure \ref{fig:community-specific-cross-taret} shows coefficients with largest differences on community-specific target ranking for 5 communities.  We observe that for most targets and communities, message passing leads to a higher correlation score. 


For target-specific community ranking, the correlation scores on ``Andrew Yang'' shows the largest improvement. Andrew Yang is known for his unique stance in the political spectrum, with policy proposals like Universal Basic Income that attracted bipartisan interest. His appeal across traditional party lines means that communities with mixed ideologies may have a more varied and nuanced view of him, which message passing can capture more effectively by incorporating a broader spectrum of opinions. In addition, Yang's campaign focused on technology, entrepreneurship, and forward-looking economic policies. These topics may resonate differently across the political spectrum, and message passing allows the model to integrate these diverse reactions better.

The underperformance of our method with message passing on the targets ``illegal immigrants'' and ``Hispanics'' may stem from the complexity and sensitivity of these issues. The topics of ``illegal immigrants'' and ``Hispanics'' are highly polarized and emotionally charged. The discussions around these subjects often involve strong opinions and biases, which can be deeply entrenched within communities. When message passing introduces opposing viewpoints or information from communities with different stances, it might not necessarily result in a more accurate representation of sentiment but could lead to a more muddled or less coherent stance that does not correlate well with the actual sentiments of individual communities.


The improvements on community-specific target ranking for Communities 5, 3, and 1 after implementing message passing, are notably more pronounced than in other communities. This observation suggests that the unique characteristics and interconnections of these specific communities make them particularly receptive to the benefits of message passing. 

Communities 5 and 3, with high percentages of liberal tweets (89\% and 88\%, respectively), both exhibit improvements in community-specific target ranking with message passing. These communities predominantly consume news from liberal sources such as Politico, Business Insider, Newsweek, The Hill, NBC News, and The Guardian. The message passing technique appears to pool nuanced liberal viewpoints from interconnected communities, enhancing the models' ability to reflect the diverse sentiments within these communities accurately.

Community 1 shows an intriguing result. Despite being the most conservative community with only 5\% liberal tweets, there is an improvement in the model with message passing. The community's top news sources, such as Fox News and Breitbart, are well-known for their conservative leanings. The introduction of message passing might be bringing in conservative but less extreme perspectives from neighboring communities, potentially offering a more nuanced representation of conservative stances. This improvement suggests that the method can refine the model's stance representation even within communities with a dominant ideological orientation by incorporating a diversity of views from within the same broader ideological spectrum.

Community 7, which predominantly shares content from liberal news outlets such as CNN, Politico.eu, The Irish Times, and The Baltimore Sun, suggests a strong liberal bias in its information dissemination. However, the inclusion of CCN, a conservative outlet, in its top-shared sources indicates some degree of ideological diversity within the community's media consumption.
Incorporating message passing into the finetuning process for community 7 could introduce more varied or even conflicting viewpoints from neighboring communities, especially if these communities share content from conservative outlets like CCN. This integration of a broader ideological spectrum could potentially dilute the community's overall liberal sentiment, leading to a less consistent and lower performance in community-specific target ranking.

\begin{table}[ht]
\addtolength{\tabcolsep}{-0.0pt}
\centering
\small
\begin{tabular}{ccc}
\Xhline{1.0pt}
            & \textbf{\begin{tabular}[c]{@{}c@{}}Finetuned GPT-2\\ +Random MP\end{tabular}} & \textbf{\begin{tabular}[c]{@{}c@{}}Finetuned GPT-2\\ + MP\end{tabular}} \\ \hline
\textbf{P1} & 45.1$\pm$1.2                                                                   & \textbf{46.7$\pm$1.4}                                                              \\
\textbf{P2} & 45.8$\pm$0.7                                                                    & \textbf{48.7$\pm$0.7}                                                              \\
\textbf{P3} & 44.2$\pm$1.3                                                                    & \textbf{48.9$\pm$1.5}                                                              \\
\textbf{P4} & \textbf{51.2$\pm$0.4}                                                                    & 49.8$\pm$0.8                                                              \\ 
\Xhline{1.0pt}
\end{tabular}
\addtolength{\tabcolsep}{0.0pt}
\caption{Spearman rank correlation of our method and an ablated method where each community exchanges information following a community retweet network whose edge weights are randomly assigned.
}
\label{tab:random-mp}
\end{table}

\noindent \textbf{Ablation Study on Random Message Passing.} A plausible counter-argument could be that the enhancement observed through our message passing approach merely results from an enlargement of each community's finetuning data pool. According to this perspective, one could just as easily enrich each corpus by drawing randomly from other community corpora, negating the need for a reference to the \emph{community retweet network}. 
In light of this, we conduct an ablation study, creating an alternative community retweet network with edge weights between communities assigned randomly. In this network the message passing does not follow the communities retweeting activities.
Comparisons between this random message passing method and our approach are illustrated in Table \ref{tab:random-mp}. Observations indicate that models finetuned with random message passing tend to underperform, providing a robust argument that our proposed method of finetuning via message passing, informed by the \emph{community retweet network}, cannot be reduced to a simplistic random data augmentation for each community's corpus. 
This further validates the crucial role played by the \emph{community retweet network} in directing the information flow and helping each community language model learn more relevant information.

\section{Conclusion}
We explore the complex ideologies of ad-hoc online communities towards different political figures and social groups. Our approach probes these ideological stances by finetuning language models on community-authored tweets and exchanging community information through message passing. Our method aligns with real-world survey data and outperforms existing baselines. Our work underscores the potential of leveraging social media data to monitor and understand societal dynamics in the digital age.

Our method offers a promising pathway for future research. Potential avenues include expanding the study to other social media platforms, analyzing how ideological stances of online communities evolve over time, and finetuning one single language model for different communities to enhance scalability when the number of communities increases. Our approach also holds the promise of providing an in-depth exploration of intricate ideological postures of the communities, facilitating a broader array of applications, including the examination of community emotional reaction to wedge issues~\cite{guo2023measuring} and affective polarization~\cite{iyengar2019origins, feldman2023affective}.

\section*{Acknowledgements}
This project has been funded, in part, by DARPA under contract HR00112290106. We appreciate the constructive advice and suggestions from the anoymous reviewers.


\section*{Limitations}

\textbf{Twitter-centric study.} Our research primarily focuses on Twitter, a single social media platform. This may limit the generalizability of our findings, as user behavior and community dynamics can vary significantly across different platforms. \\

\noindent \textbf{U.S.-centric perspectives.} We concentrate primarily on U.S. based English-speaking communities. This focus restricts the applicability of our findings, as language nuances, cultural factors, and political landscapes can greatly affect the expression and perception of ideologies in online communities.\\

\noindent \textbf{Modeling interactions through the community retweet network.} Our method relies heavily on the quality of community retweet network for information exchange. If the underlying network is not well-constructed or does not accurately reflect community interactions, it may compromise the effectiveness of our approach.\\

\noindent \textbf{Ignoring the dynamics of communities interactions.}
Our method assumes that communities are static and does not account for potential temporal changes in community formation, sentiments, interactions, and even users' political leanings.
In reality, these elements can dynamically evolve over time.\\

\noindent \textbf{Hard labeling of users' ideologies.} Following previous works \cite{rao2021political, jiang2022communitylm}, we assign binary labels to users as liberals or conservatives. However, user's political ideologies are likely to cover the full political spectrum, instead of the dichotomy of liberals and conservatives.\\

\section*{Ethics Statement}
Our study investigates online communities on Twitter, focusing on their political orientations and the propagation of different ideological stances. While this understanding is essential for addressing societal challenges such as misinformation and polarization, we are aware that our work could potentially be misused. For instance, our methods could be exploited to manipulate public opinion or target specific communities for propaganda or harassment. We condemn such misuse and advocate for the responsible application of our research findings.

Regarding data privacy, we employ publicly available Twitter data, respecting the platform's guidelines. No personal identifying information is used in our analysis, maintaining user anonymity. We acknowledge the potential risks of re-identification and take precautions to minimize this risk.

We also recognize that our work might unintentionally perpetuate biases present in the data, given that the language models are trained on real-world data, which might reflect societal biases. As such, the models' ideology probing could potentially reinforce and amplify these biases. Efforts were made to mitigate this risk by ensuring the diversity of the communities studied and clearly acknowledging this limitation in our research.

Overall, we believe that the potential benefits of our research, such as enabling better understanding of online communities and fostering healthier online discourse, outweigh these risks. However, we emphasize the need for continued ethical consideration and caution as the research progresses and its findings are put to use. 



\bibliography{anthology,custom}
\bibliographystyle{acl_natbib}


\appendix

\section{ANES Survey}
\label{app:anes}
\paragraph{30 targets studied in the ANES survey:} (1) \textit{people}: Donald Trump, Barack Obama, Joe Biden, Elizabeth Warren, Bernie Sanders, Pete Buttigieg, Kamala Harris, Amy Klobuchar, Mike Pence, Andrew Yang, Nancy Pelosi, Marco Rubio, Alexandria Ocasio-Cortez, Nikki Haley, Clarence Thomas, Dr. Anthony Fauci, and (2) \textit{groups}: blacks, whites, Hispanics, Asians, illegal immigrants, feminists, the \#MeToo movement, transgender people, socialists, capitalists, big business, labor unions, the Republican Party, the Democratic Party.

\section{Ideologies of Ad-hoc Online Communities}
\label{app:ideo_comm}
As shown in Table \ref{tab:top-comm-election}, the detected communities collectively demonstrate the diversity and variability of media consumption patterns in the online space. Each community appears to represent a unique intersection of political leanings, topical interests, and geography. For instance, some communities, such as Community 1, gravitate towards conservative news outlets, while others lean towards more liberal sources, as seen with Community 2 and 3. Another layer of differentiation comes from the specific interests or focus areas, with Community 5 showing a preference for business and Community 16 for celebrity and health-related news. Geography also play a role in news consumption, as demonstrated by outlets associated with local television news sources, like fox5ny (Community 15) and \textit{ktla} (Community 20). Overall, these differences underscore the multifaceted nature of information consumption and sharing within different communities in an online ecosystem. These observations point out the limitations of conventional methods to probe community ideologies, which rely on a predetermined binary political division \emph{left} vs \emph{right} of communities, which does not conform to the organic formalization of communities. 






\section{Community Retweet Network}
\label{app:comm-retweet-net}

\begin{figure*}[ht]
    \centering
    \includegraphics[width=0.88\textwidth]{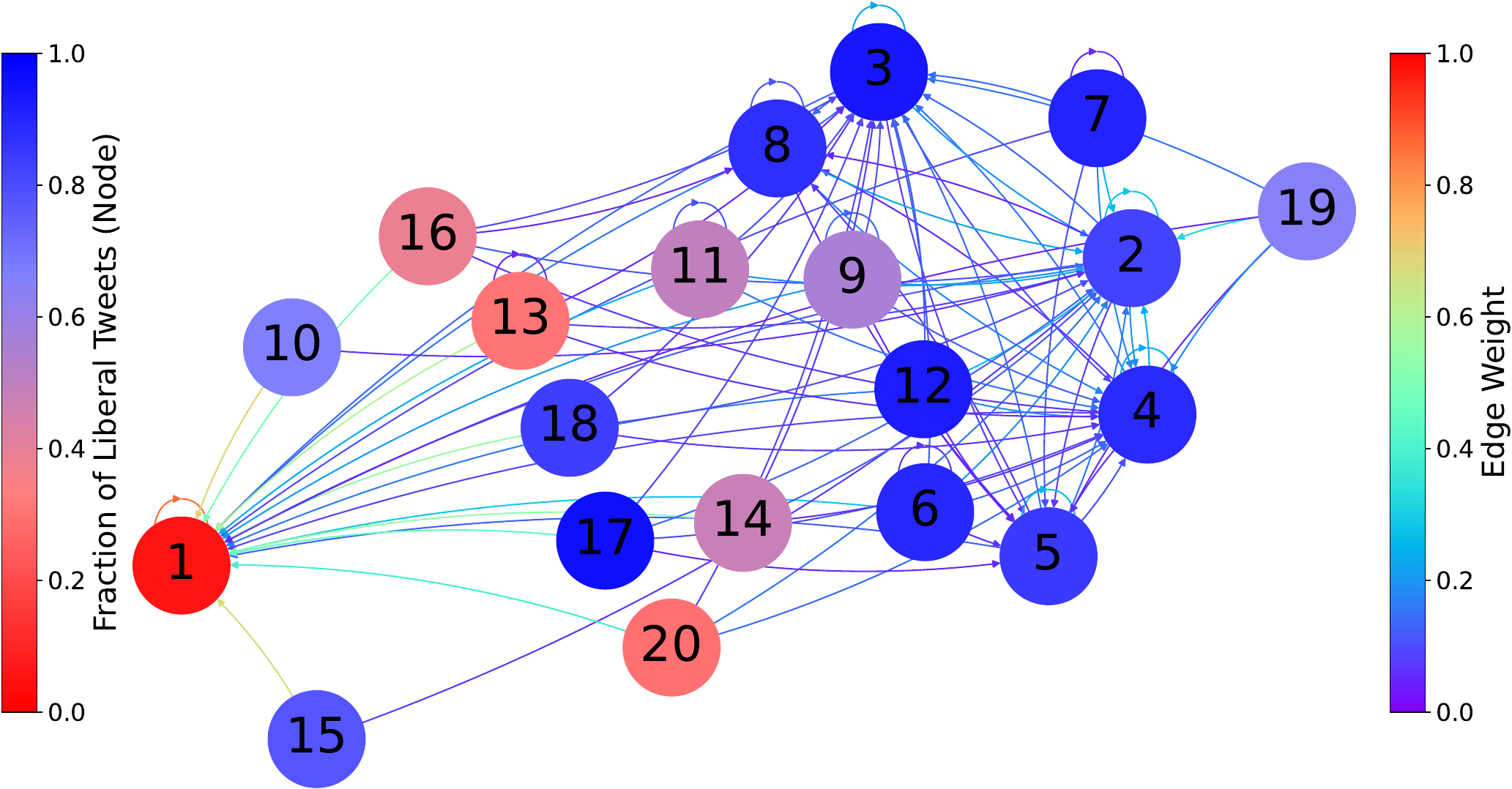}
    \caption{\emph{Community retweet network}. 
    The source node of an edge is the retweeting community, and the target node is the retweeted community. The node color represents the fraction of liberals in the community -- darker blue indicates more liberals, and darker red indicates more conservatives.
    For each community, the weights of its out edges are normalized by its out degree. Edge colors represent the edge weights.
    The edges whose weights are lower than 0.05 are not shown.}
    \label{fig:retweet-comm}
\end{figure*}

The retweet network is shown in Figure \ref{fig:retweet-comm}, where edges with weights lower than 0.05 are not shown. The node colors represent the fraction of liberal tweets in the community, and the edge colors represent the strength of connectedness between two communities.

The community retweet network illustrates the flow of information in the political discourse on social media. The darker blue nodes indicate communities with a higher fraction of liberal tweets, while darker red nodes indicate more conservative tweets. The strength of the connections, as shown by the edge colors, represents the volume of retweets between communities, revealing which communities are influential in spreading information.

Communities with many incoming edges, especially those with higher edge weights, can be considered as influential hubs within the network. These hubs are likely seen as authoritative or resonate well with the broader community, leading to their content being retweeted more frequently. For example, a community that is heavily retweeted by others may hold a significant place in shaping the discourse within its ideological alignment.

Conversely, communities with more outgoing edges are active in disseminating information, which may or may not be widely accepted or endorsed by others in the network, as indicated by the edge weights. The dynamic interplay of these retweeting patterns provides insights into how communities interact, influence each other, and contribute to the spread of ideologies across the network. This information is crucial when applying message passing techniques in finetuning language models, as it helps to understand which communities might be more receptive to certain ideologies and how they might influence the collective sentiment captured by the models.

From the retweet network we observe the following key takeaways: 1) Interconnectedness matters: The frequent retweets among communities highlight the importance of network interactions in understanding their ideologies. 2) Echo chamber phenomenon: Community 1's prevalent self-retweets (as indicated by the large weight of its self-loop) suggest a strong echo chamber effect, indicating certain conservative groups might be more ideologically isolated than their liberal counterparts. 3) Diverse news consumption: The different media outlets preferred by each community show that even communities with similar ideologies can have varied news consumption patterns, shaping their individual ideologies. 4) Comparative inclusivity of liberal communities: Communities 2 and 3 engage more with external content compared to Community 1, hinting at potentially broader information consumption.


\section{Stance Detection}
\label{app:stance_detection}
\paragraph{The reason on using sentiment analysis as a proxy of stance detection.} Admittedly, the stance towards a target expressed in a sentence might be different from the overall sentiment of the sentence, and the most ideal case would be using a pretrained stance detection \cite{he2022infusing, allaway2020zero} model on the target to detect the stance of the generated response towards it. However, not all stance detection models pretrained on the 30 targets are publicly accessible. Nevertheless, by manually inspecting the generated responses, we find that all the generated responses are simple sentences with no convoluted semantics\footnote{For example, ``Joe Biden is a joke. He is by no means presidential material.''} where sentiment analysis and stance detection would produce the same result.


To further validate this observation, for each community and target, we randomly sample 10 generated responses from our proposed finetuned GPT-2 models with message passing, and compare the sentiment labels (positive, neutral, and negative) from the sentiment analysis model to the stance labels (favor, neutral, against) towards the corresponding targets in the tweets produced by GPT-4 \cite{ouyang2022training}. We use the following prompt for inferring the stance from the generated response:

\texttt{Given the following statement and the target, infer the stance of the statement towards the target. Answer with only one word: neutral, positive, or negative.\\
Statement: [generated response]\\
Target: [target]}

By comparing the sentiment labels and the stance labels, we observe trivial (~2\%) difference between them. Therefore, it is safe to use the sentiments a proxy for the stances in our experimental setting.

\section{Experimental Setup}
\label{app:exp}
\paragraph{Model Finetuning.} We finetune the GPT-2 model on a Tesla A100 with 40GB memory. We use a batch size of 160 and learning rate of $5e-5$. We leave 2\% of data  for validation. The model is finetuned for a total of 10 epochs. When finetuning with our proposed method, message passing is conducted once after the 5th epoch, and thus every community exchanges information only with its direct neighbors.\footnote{We experimented on more frequent message passing during training, where each community could obtain information from k-hop (k$\geq1$) neighbors, but we did not see non-trivial performance improvement.} The model checkpoint with best performance (loss) on the validation set is saved for further evaluation.

\end{document}